%% file: camera_ready_main.tex
\documentclass[10pt,twocolumn,letterpaper]{article}

\usepackage{cvpr}              

\usepackage[accsupp]{axessibility}
\usepackage{algorithm2e}
\usepackage{graphicx}
\usepackage{amsmath}
\usepackage{amssymb}
\usepackage{booktabs}
\usepackage{multirow}
\usepackage{pifont}
\usepackage[T1]{fontenc}

\makeatletter
\newcommand{\setword}[2]{%
  \phantomsection
  #1\def\@currentlabel{\unexpanded{#1}}\label{#2}%
}
\makeatother

\newcommand*{\affaddr}[1]{#1} 
\newcommand*{\affmark}[1][*]{\textsuperscript{#1}}
\newcommand*{\email}[1]{\texttt{#1}}
\newcommand{\cmark}{\ding{51}}
\newcommand{\xmark}{\ding{55}}

\usepackage[dvipsnames]{xcolor}

\usepackage[pagebackref,breaklinks,colorlinks]{hyperref}

\usepackage[capitalize]{cleveref}
\crefname{section}{Sec.}{Secs.}
\Crefname{section}{Section}{Sections}
\Crefname{table}{Table}{Tables}
\crefname{table}{Tab.}{Tabs.}

\def\cvprPaperID{7033} 
\def\confName{CVPR}
\def\confYear{2023}

\begin{document}

\title{PIVOT: Prompting for Video Continual Learning}

\author{%
Andrés Villa\affmark[1,2], Juan León Alcázar\affmark[2], Motasem Alfarra\affmark[2], Kumail Alhamoud\affmark[2], Julio Hurtado\affmark[3], \\Fabian Caba Heilbron\affmark[4], Alvaro Soto\affmark[1], Bernard Ghanem\affmark[2]\\
\normalsize \affaddr{\affmark[1]Pontificia Universidad Católica de Chile}, \affaddr{\affmark[2]King Abdullah University of Science and Technology (KAUST)}, \\ \normalsize \affaddr{\affmark[3]University of Pisa, \affmark[4]Adobe Research}\\
\small \email{afvilla@uc.cl, \{juancarlo.alcazar,motasem.alfarra,kumail.hamoud\}@kaust.edu.sa} \\
\small \email{julio.hurtado@di.unipi.it, caba@adobe.com, asoto@ing.puc.cl, bernard.ghanem@kaust.edu.sa}
}
\maketitle

\input{sections/0_abstract.tex}

\input{sections/1_intro.tex}
\input{sections/2_related_works.tex}

\input{sections/3_method.tex}
\input{sections/4_results.tex}
\input{sections/5_conclusion.tex}
{\small
\bibliographystyle{ieee_fullname}
\bibliography{egbib}
}
\clearpage
\appendix
\input{sections/appendix.tex}

\end{document}

%% file: sections/0_abstract.tex
\begin{abstract}
 Modern machine learning pipelines are limited due to data availability, storage quotas, privacy regulations, and expensive annotation processes. These constraints make it difficult or impossible to train and update large-scale models on such dynamic  annotated sets. Continual learning directly approaches this problem, with the ultimate goal of devising methods where a deep neural network effectively learns relevant patterns for new (unseen) classes, without significantly altering its performance on previously learned ones. In this paper, we address the problem of continual learning for video data. We introduce PIVOT, a novel method that leverages extensive knowledge in pre-trained models from the image domain, thereby reducing the number of trainable parameters and the associated forgetting. Unlike previous methods, ours is the first approach that effectively uses prompting mechanisms for continual learning without any in-domain pre-training. Our experiments show that PIVOT improves state-of-the-art methods by a significant 27\% on the 20-task ActivityNet setup.
\end{abstract}

%% file: sections/1_intro.tex
\section{Introduction}
\label{sec:intro}

Modern Deep Neural Networks~(DNNs) are at the core of many state-of-the-art methods in machine vision \cite{he2016deep, tan2019efficientnet, ren2015faster, he2017mask, vaswani2017attention, devlin2018bert} tasks. To achieve their remarkable performance, most DNNs rely on large-scale pre-training \cite{krizhevsky2017imagenet, vaswani2017attention, dosovitskiy2020image, clip}, thereby enabling feature reuse in related downstream tasks \cite{argyriou2006multi}. However, adapting and fine-tuning a pre-trained DNN on a novel dataset commonly leads to \textit{catastrophic forgetting} \cite{french1999catastrophic}. This phenomenon explains how the effectiveness of the fine-tuned DNNs  drastically reduces in the original training distribution, in favor of increased performance on the downstream task.

This undesirable property has driven multiple research efforts in the area of Continual Learning (CL) \cite{gem,agem,chaudhry2019tiny,ewc,ghunaim2023real}, yielding techniques that enable fine-tuning a DNN on a sequence of tasks while mitigating the performance drop along the intermediate steps. One of the most challenging scenarios for the study of CL is Class Incremental Learning (CIL), where the labels and data are mutually exclusive between tasks, training data is available only for the current task, and there are no task identifiers on the validation step (\textit{i.e} task boundaries are not available at test time). Such a setup requires learning one model that, despite the continuous adaptation to novel tasks, performs well on all the seen classes.
\begin{figure}
    \centering
    \includegraphics[width=0.8\linewidth, trim={0 0.5cm 0 0}]{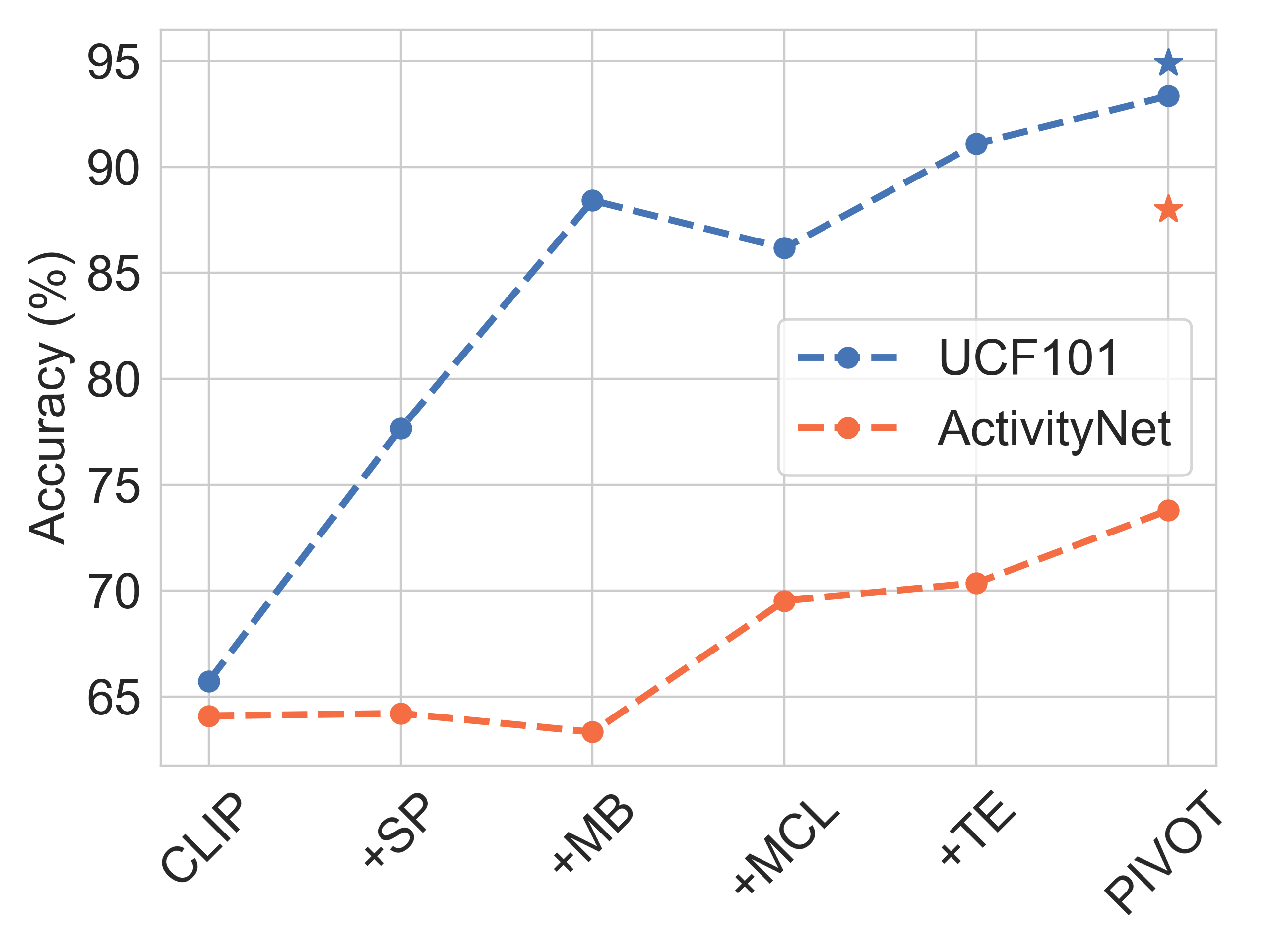}\vspace{-0.2cm}
    \caption{\textbf{Performance improvement by each PIVOT component.} We report the average accuracy on all tasks under the 10-task CIL on UCF101 and ActivityNet. We report the performance of basic CLIP, and then gradually equip it with other components: \textbf{S}patial \textbf{P}rompting, \textbf{M}emory \textbf{B}uffer, \textbf{M}ulti-modal \textbf{C}ontrastive \textbf{L}earning, \textbf{T}emporal \textbf{E}ncoder, to finally reach our proposed PIVOT method. The addition of each proposed component generally boosts the performance on both datasets. 
    Stars represent the upper bound performance on each benchmark.
    }\vspace{-0.3cm}
    \label{fig:my_label}
\end{figure}
Recent advances in mitigating catastrophic forgetting rely on deploying episodic memory or regularization techniques \cite{rebuffi2017icarl, chaudhry2019tiny, ewc, mas}. Nevertheless, most of this progress has been directed toward analyzing the catastrophic forgetting of DNNs in the image domain. The works of \cite{pellegrini2020latent, Park_2021_ICCV, Villa_2022_CVPR} introduced the CIL setup to the video domain, in particular the action recognition task. Unlike its image counterpart, video CIL requires careful modeling of the temporal information, making it an even more challenging setup. Despite the success in small-scale datasets (like UCF101), state-of-the-art methods have shown limited effectiveness on more challenging video test-beds built upon larger action taxonomies, such as Kinetics and ActivityNet \cite{Villa_2022_CVPR}.

Currently, state-of-the-art methods for video CIL rely on temporal masking and feature distillation to mitigate catastrophic forgetting\cite{Park_2021_ICCV, Villa_2022_CVPR}. In this paper, we take inspiration from recent advances in large-scale DNNs for zero-shot image classification\cite{clip} and learnable prompts for continual learning \cite{Wang_2022_CVPR}, and we propose a novel strategy for CIL in the video domain. We show that a zero-shot baseline pre-trained in the image domain already outperforms the best CIL methods in the action recognition task\footnote{A similar result for image domain CIL is outlined in the concurrent work Thengane \textit{et al.} \cite{thengane2022clip}}. Moreover, we show that this baseline can be significantly improved by enabling temporal reasoning and augmenting the modality encoders with a novel prompting strategy. As a consequence, our proposed method, PIVOT, outperforms every other baseline, setting a new state-of-the-art in all the 3 datasets included in the challenging vCLIMB benchmark for video CIL \cite{Villa_2022_CVPR}. Figure \ref{fig:my_label} summarizes the performance improvements of our approach.

Notably and following the core ideas of the vCLIMB \cite{Villa_2022_CVPR} benchmark, PIVOT does not rely on any in-distribution pre-training (a common feature of prompting methods for CL \cite{Wang_2022_CVPR, thengane2022clip}). Rather, it leverages the vast and general visual knowledge contained in the CLIP visual encoder (trained on massive amounts of paired static images and text) and maps that knowledge into a feature space suitable for video understanding in a continual learning setup.



\textbf{Contributions.} This paper proposes PIVOT (PromptIng for Video cOnTinual learning), a novel strategy for continual learning in the video domains that leverages large-scale pre-trained networks in the image domain. Our work brings the following contributions: \textbf{(i)} We show that a multimodal classifier (Video-Text) mitigates catastrophic forgetting while greatly increasing the final average CIL accuracy. \textbf{(ii)} We design the first prompt-based strategy for video CIL. Our approach leverages image pre-training to significantly mitigate forgetting when learning a sequence of video action recognition tasks.
\textbf{(iii)} 
We conduct extensive experimental analysis to demonstrate the effectiveness of PIVOT.
Our results show that PIVOT outperforms state-of-the-art methods in the challenging vCLIMB benchmark by 31\%, 27\%, and 17.2\% in the 20-task setups of Kinetics, ActivityNet, and UCF101, respectively.

%% file: sections/2_related_works.tex
\section{Related Work}

\paragraph{Image Continual Learning.} Different approaches to tackle Continual Learning have been proposed over the last few years. These methods can broadly be classified into three categories: regularization, memory, and parameter isolation methods \cite{delange2021clsurvey}. Regularization-based approaches penalize changes to weights associated with previous tasks. Among these methods, there is a first sub-group that attempted to minimize the differences in weights that are relevant to previous tasks. There are several techniques to determine the relevance of each weight, including the Fisher matrix \cite{ewc}, the gradient \cite{zenke2017continual, mas}, the uncertainty \cite{ebrahimi2019uncertainty}, among others \cite{aljundi2018selfless, saha2021gradient}. An alternative to previous approaches is to use distillation over previous task representations while training on the new one \cite{lwf, smith2021always, gao2022rdfcil}. While regularization methods can mitigate forgetting, they usually under-perform in the video domain, as their extension to temporal data remains an open research topic \cite{Villa_2022_CVPR}.

Memory-based methods store a subset of training samples from past tasks to be used in the future as replay \cite{chaudhry2019tiny, ebrahimi2021remembering} or regularization \cite{gem, agem}. Alternatively, some methods train generative models to approximate the image distribution of past tasks. Synthetic images drawn from the generative model are then used as replay \cite{shin2017continual, lesort2019generative, hayes2020remind, simcs}. While showing high effectiveness in the image domain, these methods are not directly applicable to the video domain for two main reasons. First, due to the higher memory usage when storing video samples and high complexity of generating videos. Second, due to the associated challenge of selecting the key video segments to store in memory \cite{Villa_2022_CVPR}. 

On the other hand, parameter-isolation methods aim to overcome the distribution shift by learning isolated parameters for each task \cite{hu2019overcoming}. While some of these methods clone the model when a new task arrives \cite{rusu2016progressive, fernando2017pathnet}, others prefer to use masks \cite{mallya2018piggyback, wortsman2020supermasks} or conditioning a trainable knowledge base \cite{ebrahimi2020adversarial, hurtado2021optimizing} to select relevant information for the current input. A critical problem with these approaches is identifying the corresponding weights for each input. Given that only a sub-group of weights is specialized for a task, the task-id is needed during training and validation to identify the corresponding group.

\vspace{2pt}\noindent\textbf{Adapting General Visual Knowledge.} Instead of continually learning reusable representations \cite{hurtado2021optimizing, mendez2022reuse}, some approaches use fixed large-scale pre-trained models as knowledge bases \cite{mehta2021empirical, Wang_2022_CVPR, thengane2022clip}. The large amount of data used to train these models helps them generalize to multiple concepts, making it possible to apply zero-shot classification for continual learning \cite{thengane2022clip} without fine-tuning. Other works have augmented these pre-trained models with learnable prompts \cite{Wang_2022_CVPR, wang2022dualprompt}, extracting better representations, significantly improving performance, and reducing forgetting in image CIL. However, these methods do not consider the temporal nature of videos, so they are not directly applicable. Although some works \cite{PanEtAl, VideoCLIP} have explored transferring image domain knowledge of \cite{thengane2022clip} to video domain, \cite{Castro_2022_BMVC} has demonstrated that CLIP pre-training on image-text pairs is more transferable than a pre-training on noisy video-text pairs. In this work, we take a first step in leveraging large-scale pre-trained networks, \ie CLIP, and equip them with temporal reasoning and learnable prompts to mitigate forgetting in video CIL.

\vspace{2pt}\noindent\textbf{Video Continual Learning.} Only a few works have addressed the unique challenges of video continual learning by proposing video-specific strategies to mitigate forgetting. In the case of \cite{Park_2021_ICCV}, the authors proposed to encode the temporal information through temporal-based knowledge distillation. A second work \cite{Villa_2022_CVPR} resolves the complexity of storing videos in memory by employing a temporal consistency loss, which reduces the number of frames stored. Despite the progress, such methods do not scale well on real-world action recognition benchmarks such as Kinetics \cite{kay2017kinetics} and ActivityNet \cite{caba2015activitynet}. In this paper, we extend the ideas presented in previous works and propose PIVOT, which exploits an image-language  pre-trained model (CLIP) and extends it temporally. PIVOT enables significant performance gains on both small and large-scale video CIL benchmarks.


%% file: sections/3_method.tex
\section{Prompting for Video Continual Learning}\label{sec:method}

\begin{figure*}[t]
    \centering
    \includegraphics[width=0.9\linewidth]{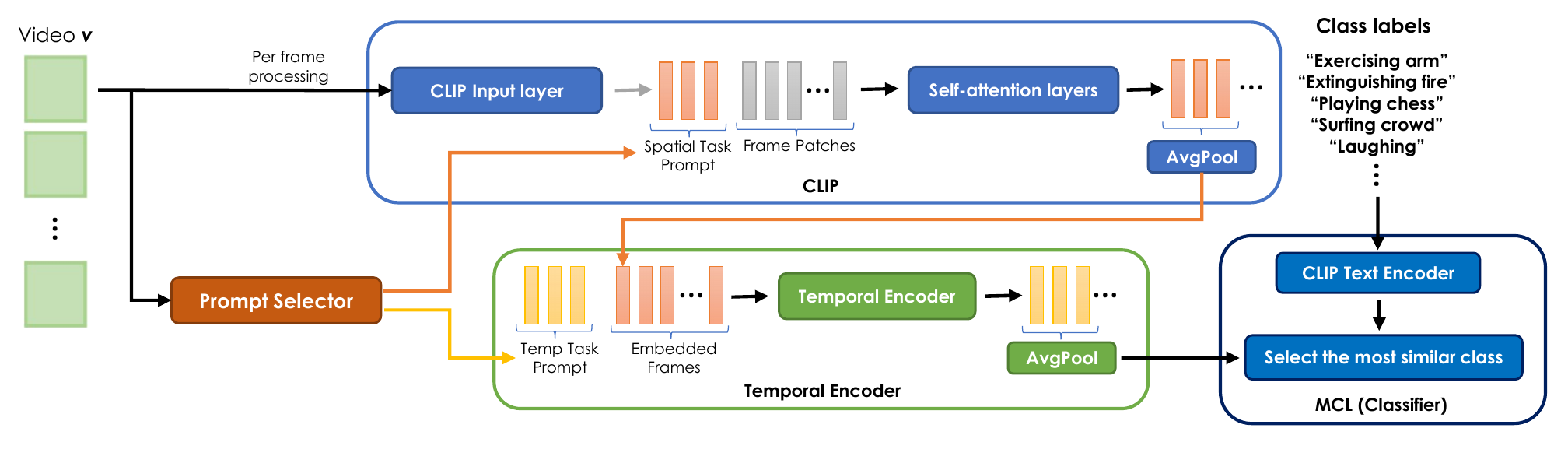}
    \caption{
    \textbf{Outline of the PIVOT Model at inference time.} Our model follows 3 main stages to make a prediction in video CIL setup: \textbf{(1)} PIVOT selects the specific task prompts for a video instance based on its similarity with the keys $K$ of each task prompt (Blue box). \textbf{(2)} Then, PIVOT adds the selected spatial and temporal task prompts to the corresponding encoder spatial (CLIP) or Temporal Encoder (green box). \textbf{(3)} Finally, PIVOT classifies the video following an MCL, considering the similarity with text embeddings from the labels of all the available classes computed with the CLIP text encoder (black box).
    }
    \vspace{-4mm}
    \label{fig:model_fig}
\end{figure*}


Our main objective is to leverage the knowledge available in large-scale image-language models for video CIL. That is, we want to learn how to adapt large visual-language DNNs available in the image domain into a single end-to-end network that is effective in video CIL. Furthermore, we want to make sure that such a transfer of knowledge involves no fine-tuning of the image encoder, and requires minimal extensions over the original network. Such constraints will ultimately favor less forgetting in the CIL setup, as the key parameters will suffer less drift during fine-tuning on different tasks.

Our approach, PIVOT, is divided into three major components.
\textbf{(i)} We adapt a large-scale image-language pre-trained model, CLIP \cite{clip}, to extract visual information from video data.
\textbf{(ii)} We aggregate the temporal information extracted from several frames within a video via a transformer encoder.
\textbf{(iii)} We reinforce the learned knowledge from each task by deploying learnable tokens, \ie prompts, to prevent the forgetting of previously learned tasks.
Despite its multiple components, we train PIVOT in an end-to-end fashion. We highlight that this training process is lightweight since the image and text encoders are frozen.
Before we delve into the details of each of our model components, we give a brief preliminary on the Video Class Incremental Learning~(VCIL) problem and the multi-modal contrastive classifier proposed in CLIP, as it plays a key role in our approach. 


\subsection{Problem Definition}
We follow the recent vCLIMB benchmark~\cite{Villa_2022_CVPR} training a single model $f_\theta: \mathcal X \rightarrow \mathcal Y$, \ie a DNN parameterized with $\theta$, that learns to predict an action label $y\in\mathcal{Y}$ given an input video $x\in\mathcal{X}$. The model learns from a sequence of tasks $\{(X_1, Y_1),(X_2, Y_2),\dots,(X_n, Y_n)\}$ where $X_i\subseteq\mathcal X$ and $Y_i \subseteq \mathcal Y \,\,\forall i$.
In the class-incremental setup, we assume that different tasks present the model with a different set of labels, \ie $Y_i \cap Y_j = \emptyset \,\,\forall i\neq j$.
We define $Acc_i$ as the average classification accuracy of all the tasks up to and including task $i$. Our objective is to train a single model $f_\theta$ that maximizes the average accuracy $Acc$, which is the average over all observed tasks.

\subsection{Multi-modal Contrastive Classifier for Video Continual Learning}\label{sec:mcl}

Our method builds on top of the image-language pre-trained CLIP encoder. CLIP \cite{clip} demonstrated that textual information promotes the learning of transferable visual features. PIVOT re-purposes the CLIP encoder for video CIL by using its multi-modal contrastive classifier as a video feature generator. We use CLIP's language encoder to extract text embeddings from the labels of all the available classes, and  its image encoder to extract visual embeddings for arbitrary frame data. These paired embeddings allow us to perform a pseudo-classification by looking for tuples that minimize the distance between visual and textual embeddings. By keeping the CLIP's encoder frozen and forwarding video related data (in the visual and language streams), we allow the transfer of its extensive knowledge into the video domain in a zero-shot fashion. 

We observe three advantages in Multi-modal Contrastive Learning~(MCL) for the continual learning setup.
First, this approach is parameter-free. Hence, our classifier would not suffer from \emph{any} forgetting when learning different tasks.
This mitigates the need to deploy a large replay buffer to maintain the performance of the classifier on previously learned tasks.
Second, this approach directly leverages the frozen CLIP encoder in the video CIL setup.
We empirically find that, despite the simplicity of this approach and the domain gap between image and video data, a zero-shot CLIP significantly outperforms all continual learning methods in the literature on both Kinetics and ActivityNet VCIL setups. Third, we find the core idea of the zero-shot approach (multi-modal contrastive loss) could be used to augment the zero-shot strategy, thus enabling prompting mechanisms, temporal modeling, and replay memories that still rely on a frozen CLIP encoder. 



\subsection{PIVOT Model}\label{sec:pivot_model}

Prompting methods \cite{Wang_2022_CVPR, wang2022dualprompt} have emerged as an alternative to traditional rehearsal methods. While prompting mechanisms have been shown effective in drastically reducing memory sizes, they rely on in-distribution pre-training to make suitable prompt selections \cite{Wang_2022_CVPR}. We approach a more general case, where we learn to prompt for video data while relying exclusively on image-language knowledge bases. To this end, we propose PIVOT, a Prompting-based Model for Video Continual Learning. PIVOT leverages CLIP $(f_{sp})$ to encode spatial information and a transformer as temporal encoder $(f_{tp})$ to model the temporal relations essential to understanding actions. Additionally, we define a novel spatial and temporal prompting scheme that captures specific properties of each task to increase the model performance while mitigating the forgetting of previously learned tasks. 

We outline a 3-stage process to effectively train and select prompts for the spatial and temporal dimensions of videos by leveraging extensive image domain pre-training. Despite having no video domain pre-training, our prompting strategy is effective on the action recognition task. To the best of our knowledge, this is the first prompting method that can work in the presence of such a domain shift. 

\vspace{2pt}\noindent\textbf{Out-of-distribution Adaptation.} The first training stage adapts an image pre-trained model to extract video features. Given an input video $v \in \mathbb{R}^{T \times H \times W \times C}$, where $T$ is the number of video frames, we utilize CLIP to encode the video frames into meaningful spatial features $ v_{sp} = f_{sp}(v) \in \mathbb{R}^{T \times D_m}$. 
These spatial features are then passed through the temporal encoder where we append a $[class]$ token that aggregates the temporal information of the feature sequence, resulting in the temporal feature of the video $ v_{tp} = f_{tp}(v_{sp})[class] \in \mathbb{R}^{D_m}$. During this process, the CLIP encoder is completely frozen to avoid any distribution shift. In contrast, the temporal encoder is trained with the current task data $(X_i, Y_i)$. Specifically, we train the temporal encoder following the MCL approach outlined in Section~\ref{sec:mcl}. By using an approach that is informed by the CLIP frozen features, we ensure that the learned features of the temporal model stay consistent with the knowledge of the original image-language model. 

\paragraph{Task-specific Prompt Generation.} After tuning the temporal component of our model, the second training stage aims at learning task-specific prompts. Current prompting methods \cite{Wang_2022_CVPR} would rely on continuously updating a prompt pool along the full task set. Since we can not rely on in-distribution pre-training to update and select prompts, we adopt task-specific prompts that preserve individual task knowledge along the CL process. Essentially, we train a new set of prompts as the data for a new task becomes available, then we append these new prompts into the pre-existent pool. 


In addition to maintaining task-specific prompts, we propose to train the prompts and the prompt selector separately. At training time, we optimize the task-specific prompts $P_{n}$ for task $n$ following the MCL approach, while the prompts associated to previous tasks ($P_{n-1} \cup P_{n-2} \cup ... P_{0}$) remain fixed. To address the temporal nature of video data, each task will estimate independent spatial and temporal prompts $P_n = (P_n^{sp}, P_n^{tp})$, where $P_n^{sp} \in \mathbb{R}^{N_p \times L_{sp} \times D_{in}}$ and $P_n^{tp} \in \mathbb{R}^{N_p \times L_{tp} \times D_m}$, $N_p$ is the number of prompts per task,  $L_{sp}$ and $L_{tp}$ are the lengths of spatial and temporal prompts, respectively, and $D_{in}$ is the output dimension of the CLIP input layer $f^e_{sp}$.

\vspace{2pt}\noindent\textbf{Prompt Selection}\label{par:prompts}
Unlike L2P \cite{wang2022dualprompt}, we can not use extensive in-domain knowledge to select the best prompts for each input. To alleviate CLIP's lack of domain comprehension, and thus improve its selection performance, we train the temporal encoder to select the prompts. We first associate each task-specific prompt with the feature set  estimated from the frozen CLIP representations. To this end, we use the feature estimated on the language stream of CLIP, using the labels as textual input to the stream, this feature set will define the key $K_{n}$ to index a prompt: $\{(K_1, P_1), (K_2, P_2),\dots,(K_n, P_n)\}$, where $K_n \in \mathbb{R}^{M_n \times D_m}$, $M_n$ is the number classes on task $n$, and $D_m$ is the output dimension of CLIP Text encoder. 

We train the temporal encoder to select the corresponding task prompts for each input following a pseudo-classification based on the distance estimated by the MCL. We select the task prompts with the most similar key to the final video representation $v_{tp}$. Likewise, once the model selects the prompts, we train the temporal encoder to support them with an MCL.
Figure \ref{fig:model_fig} illustrates the PIVOT pipeline.




\subsection{PIVOT Forward Pass }\label{sec:classifying_with_pivot}
In Section~\ref{sec:pivot_model}, we outlined the construction and selection of both spatial and temporal prompts. 
Next, we discuss how do we use these prompts during inference.

First, we note that CLIP $f_{sp}$ is based on the ViT architecture, so, it is composed of input layer $f^e_{sp}$ and self-attention layers $f^s_{sp}$. 
Given an input video $v$, we pass it through $f^e_{sp}$, which divides each frame into $L$ patches and encodes them to $v_{sp} = f^e_{sp}(v) \in \mathbb{R}^{T \times L \times D_{in}}$. We leverage the spatial prompts at this stage, and concatenate $P_n^{sp}$ with $v_{sp}$ resulting in $v^{e}_{sp}=[P_n^{sp}; v_{sp}] \in \mathbb{R}^{T \times (N_pL_{sp}+L) \times D_{in}}$ and pass the extended sequence to the self-attention layers $v^{s}_{sp} = f^s_{sp}(v^{e}_{sp}) \in \mathbb{R}^{T \times (N_pL_{sp}+L) \times D_{m}}$. Following \cite{Wang_2022_CVPR}, for the final spatial representation we take an average pooling (AvgPool) through the first $N_pL_{sp}$ tokens of $v^{s}_{sp}$ resulting in $v^{avg}_{sp}\in\mathbb R^{T\times D_m}$. It is important to note that, we use the same spatial prompt for all frames in a given video.


The previous process results in an embedding vector for each frame that takes into account both the extracted knowledge from CLIP and the learnt spatial prompts.
We then aim at aggregating the information across all frames by leveraging a temporal encoder $f_{tp}$.
Before we do so, we employ the learnt temporal prompts in a similar fashion to the spatial ones.
In particular, we first concatenate $x = [P_n^{tp}; v^{avg}_{sp}]$ and then pass it through the temporal encoder followed by an average pooling layer. Formally $v_{tp} = AvgPool(f_{tp}(x)[:N_pL_{tp}])  \in \mathbb{R}^{ D_{m}}$.
At last, we use $v_{tp}$ as a global feature representation for the input video to classify it with the MCL classifier outlined in Section~\ref{sec:mcl}. See Section~\ref{sec:supp_prompt_selector} in supplementary material for more details.

\subsection{Training PIVOT}\label{sec:training_prompts}
Our training pipeline is divided into three stages for each task in the VCIL setup.
We first train our temporal encoder to correctly classify input data within the current task.
Then, we leverage the knowledge of both CLIP and the trained temporal encoder to learn per task learnable tokens, \ie prompts.
Finally, we fine tune the temporal encoder to select and leverage the learned prompts to enhance the performance and reduce forgetting.
It is worth mentioning that we freeze CLIP throughout the entire training process. 

\vspace{2pt}\noindent\textbf{First Stage: Out-of-distribution Adaptation.} 
In this stage, we want to align the temporal encoder with the extracted representation of a given video from CLIP.
To do so, we train our temporal encoder to classify a video based on the features extracted from CLIP with an MCL loss.
Let $\theta_{tp}$ be the learnable parameters for $f_{tp}$, we seek to minimize the loss for the $n^{th}$ task:
\begin{equation}\label{eq:first_stage_training}
\min_{\theta_{tp}} \mathbb E_{(v, y)\sim (X_n, Y_n)}\left[ \mathcal{L}\left(f_{tp}(f_{sp}(v)), y\right)\right],
\end{equation}
where y is the representation of the class label of v obtained with the CLIP text encoder, $\mathcal L$ is the MCL loss, and $(X_n, Y_n)$ is the data distribution for the $n^{th}$ task.
We note here that relying exclusively on Eq.~\eqref{eq:first_stage_training} for training might result in forgetting previously learnt tasks.

\vspace{2pt}\noindent\textbf{Second Stage: Task-specific Prompt Generation.} 
In this stage, we leverage the knowledge encoded on the temporal encoder to learn meaningful prompts which are specific for task $n$. To that end, we freeze the temporal encoder during this stage. Then we randomly initialize a set of prompts for task $n$ denoted as $P_n = (P_n^{sp}, P_n^{tp})$.
We include $P_n$ in the forward propagation of our pipeline as discussed in Section~\ref{sec:classifying_with_pivot}.
We then learn the prompts $P_n$ by minimizing the following loss:
\begin{equation}\label{eq:learning_prompts}
\min_{P_{sp}, P_{tp}}  \mathbb E_{(v, y)\sim (X_n, Y_n)}\left[\mathcal{L}\left(f_{tp}(f_{sp}(v, P^{sp}_n),P^{tp}_n), y\right)\right].
\end{equation}

Note that Eq.~\eqref{eq:learning_prompts} aims at aligning the learnable prompts with the knowledge of both CLIP and the learnt temporal encoder for the same task.
This is accomplished by the prompts $P_n$.
We note here that we freeze the prompts for a given task while learning prompts for other tasks.

\vspace{2pt}\noindent\textbf{Third Stage: Prompt Selection.}
In the last stage, we fine-tune the temporal encoder over the replay memory data $M$, $M$ includes data samples for the current task.
This is accomplished through minimizing the combined MCL loss from Equations~\ref{eq:first_stage_training} and \ref{eq:learning_prompts}:
\begin{equation}
\begin{split}
\min_{\theta_{tp}} \mathbb E_{(v, y)\sim M}\big[ \mathcal{L}(f_{tp}(f_{sp}(v, P_{sp}),P_{tp}), y) \\ +  \mathcal{L}(f_{tp}(f_{sp}(v)), y) \big].
\end{split}
\label{eq:last_step}
\end{equation}

It is worth mentioning that the last training step in Eq.~\eqref{eq:last_step} aligns the parameters of the temporal encoder to solidify the knowledge of a given task. This is achieved by training over the selected prompts and the network's  input. 
This will reduce forgetting previously learnt tasks as their corresponding prompts are frozen.

%% file: sections/4_results.tex
\section{Experiments}
\label{sec:results}

\begin{table*}[t!]
\caption{
\textbf{Results on the two most challenging datasets of vCLIMB}. We follow the vCLIMB Benchmark \cite{Villa_2022_CVPR} and report the average accuracy (Acc) and the backward forgetting (BWF) at 10 and 20 tasks. We highlight that CLIP zero-shot \cite{clip} outperforms regularized-based methods such as EWC \cite{ewc} and MAS \cite{mas} and Memory-based methods such as iCaRL \cite{rebuffi2017icarl} and BiC \cite{Wu_2019_CVPR}. Our approach PIVOT, which builds on top of CLIP, outperforms all the previous methods, by up to $31\%$ in the Kinetics and ActivityNet 20-task sequence.
}\vspace{-1mm}
\centering
 \small
\resizebox{17cm}{!}{
\begin{tabular}{ccccccccccc} 
    \toprule

    \multicolumn{1}{c}{\multirow{4}{*}{\bf Model}} 
    & \multicolumn{5}{c}{\bf Kinetics} & \multicolumn{5}{c}{\bf ActivityNet-Trim}
    \\\cmidrule(lr){2-11}
         & \multicolumn{1}{c}{\multirow{2}{*}{ \begin{tabular}[c]{@{}c@{}}\textbf{Mem. Video} \\ \textbf{Instances} \end{tabular}}} & \multicolumn{2}{c}{\bf 10 Tasks} & \multicolumn{2}{c}{\bf 20 Tasks} & \multicolumn{1}{c}{\multirow{2}{*}{ \begin{tabular}[c]{@{}c@{}}\textbf{Mem. Video} \\ \textbf{Instances} \end{tabular}}} & \multicolumn{2}{c}{\bf 10 Tasks} & \multicolumn{2}{c}{\bf 20 Tasks}
         \\\cmidrule(lr){3-6}\cmidrule(lr){8-11}
         & & \multicolumn{1}{c}{\bf Acc $\uparrow$ } & \multicolumn{1}{c}{\bf BWF $\downarrow$ } & \multicolumn{1}{c}{\bf Acc $\uparrow$ } & \multicolumn{1}{c}{\bf BWF $\downarrow$ } & & \multicolumn{1}{c}{\bf Acc $\uparrow$ } & \multicolumn{1}{c}{\bf BWF $\downarrow$ } & \multicolumn{1}{c}{\bf Acc $\uparrow$ } & \multicolumn{1}{c}{\bf BWF $\downarrow$ } 
         \\
    \midrule
    EWC & None & 5.81\% & 16.05\% & 2.95\% & 32.70\% & None & 4.02\% & 5.32\% & 1.28\% & 3.77\%
    \\
    MAS & None & 7.81\% & 10.12\% & 4.25\% & 5.54\% & None & 8.11\% & 0.18\% & 4.61\% & 0.1\%
    \\
    \cmidrule{1-11}
    BiC & 8000 & 27.90\% & 51.96\% & 23.06\% & 58.97\% & 4000 & 51.96\% & 24.27\% & 46.53\% & 15.95\% 
    \\
    iCaRL & 8000 & 32.04\% & 38.74\% & 26.73\% & 42.25\% & 4000 & 48.53\% & 19.72\% & 43.33\% & 21.57\%
    \\
    \cmidrule{1-11}
    CLIP & None & 46.50\% & 9.80\% & 46.52\% & 11.32\% & None & 64.1\% & 10.47\% & 64.13\% & 11.71\%
    \\
    \cmidrule{1-11}
    PIVOT & 4000 & 55.13\% & 26.50\% & 55.04\% & 26.41\% & 2000 & \bf 73.8\% & \bf 11.41\% & \bf 73.84\% & \bf 11.94\%
    \\
     PIVOT w/o prompts & 4000 & \bf 58.61\% & \bf 20.78\% & \bf 57.51\% & \bf 21.98\% & 2000 & 72.22\% & 13.71\% & 73.43\% & 13.45\%
    \\

    \cmidrule{1-11}
    Upper bound & -- & 73.9\% & -- & 73.9\% & -- & 2000 & 88\% & -- & 88\% & --
    \\
    \bottomrule
    \end{tabular}
}
\vspace{-3mm}
\label{tab:main_results}
\end{table*}

We perform extensive experiments to benchmark existing methods against PIVOT in the VCIL action recognition task. In addition to comparing PIVOT directly to the existing methods reported in vCLIMB, we benchmark the performance of our frozen CLIP baseline. Then, we conduct thorough ablation studies to understand the role that each component of PIVOT contributes to its final performance. See our supplementary material for additional experiments.

\subsection{vCLIMB Benchmark}
We follow the vCLIMB benchmark~\cite{Villa_2022_CVPR} and conduct our experiments on Kinetics~\cite{kay2017kinetics}, ActivityNet~\cite{caba2015activitynet}, and UCF101~\cite{soomro2012ucf101} datasets where classes of each dataset are presented sequentially over 10 or 20 tasks. For example, in the 10-task schedule of ActivityNet, the 200 classes are broken into 10 sets, each containing 20 non-overlapping categories. We report the average accuracy (Acc) of $f_\theta$ on all observed tasks ($n$), where higher values indicate better classification performance. Additionally, we follow the standard practice in VCIL~\cite{Villa_2022_CVPR}, and report the backward forgetting~(BWF), where smaller values indicate less forgetting and hence are better. Equation \ref{eq:metrics} shows the formulas for the metrics, where $A_{i,j}$ is the accuracy of task $j$ after training task $i$.  
\begin{equation}
    \begin{split}
        Acc &= \frac{1}{n} \sum_{i=1}^n A_{n,i} \\
        BWF &= \frac{1}{n-1} \sum_{i=1}^{n-1} A_{i,i} - A_{n,i} 
    \end{split}
    \label{eq:metrics}
\end{equation}
\paragraph{Implementation details.} 
PIVOT is augmented with task-specific prompts (see Algorithm~\ref{alg:pivot_v2_forward_pass} in the supplementary material). We add $N_p=1$ set of spatial and temporal prompts per task with lengths $L_{sp}=3$ and $L_{tp}=3$, respectively. Therefore, at the end of the CL training process, PIVOT has 10 and 20 sets of spatial and temporal prompts in the 10-tasks and 20-tasks scenarios, respectively. Hence, PIVOT incorporates all three stages outlined in \ref{sec:pivot_model}, each stage taking 40 epochs using SGD optimizer with a constant learning rate of 0.01 and a batch size of 50. The PIVOT temporal encoder is a transformer encoder with 2 heads and 3 layers. We use the CLIP model for all experiments, employing a ViT-B/32 architecture. $T = 8$ frames per video are sampled based on the strategy proposed in \cite{TSN}. Moreover, we define "PIVOT without prompts" (see Algorithm~\ref{alg:pivot_forward_pass} in the supplementary material). It uses the very same architecture as PIVOT but lacks any prompting mechanisms. It is trained for 40 epochs using the same optimizer, learning rate, and batch size as PIVOT, following the MCL approach on every CIL task. It's worth noting that all methods with a temporal encoder in our ablation study (see Section~\ref{sec:ablation_study}) follow the same optimization setup as PIVOT w/o prompts.

On the contrary, the augmented CLIP model with spatial prompts follows the prompting approach of \cite{Wang_2022_CVPR}. Therefore, it follows a similar setup as \cite{Wang_2022_CVPR}: an Adam optimizer with a learning rate of 0.03, $\beta_1 = 0.9$, $\beta_2 = 0.999$, and a batch size of 50. For this spatial prompting, we consider 10 prompts in total, each with a length $L_{sp} = 5$, and select 5 prompts per video instance using the same selection approach as \cite{Wang_2022_CVPR}.

\subsection{Comparison with Baselines.}
We compare the performance of our proposed methods PIVOT and PIVOT w/o prompts, and our CLIP baseline, against the approaches reported in the recent video class incremental learning benchmark vCLIMB~\cite{Villa_2022_CVPR}, namely EWC, MAS, BiC, and iCaRL. PIVOT differs from PIVOT w/o prompts by employing task-specific prompts while sequentially learning the video tasks. We further report the upper bound performance that is acquired by training the TSN backbone~\cite{TSN} offline, i.e., on all the dataset classes simultaneously. 
The results for Kinetics and ActivityNet are shown in Table \ref{tab:main_results}, while the results for UCF101 are reported in Table~\ref{tab:ucf_results}. Note that some methods store past-task exemplars to mitigate forgetting. We set the memory budget for these replay-based methods proportionally to the amount of classes of the dataset. The number of video instances used in each setup is stated in the respective table. 

\vspace{2pt}\noindent\textbf{SOTA Comparisons to PIVOT.}
Our results show that PIVOT and PIVOT w/o prompts outperform all the methods in vCLIMB by large margins. We first highlight that the highest performing method in vCLIMB was iCaRL, which had an average accuracy of $26.73\%$ and $43.33\%$ on 20-task Kinetics and 20-task ActivityNet respectively. The average accuracy of $57.51\%$ obtained by PIVOT w/o prompts is $31\%$ higher than iCaRL in the Kinetics 20 task set. Moreover, using PIVOT results in a remarkable improvement of $30\%$ over the best result reported in vCLIMB for the 20 task set on the ActivityNet dataset. Regardless of the dataset used, our approach leverages image-pretraining, and a multi-modal contrastive learning approach to consistently outperform the prior VCIL art. 

\vspace{2pt}\noindent\textbf{Image CLIP is a strong VCIL baseline.} Following \cite{thengane2022clip}, we also propose to use a frozen CLIP~\cite{clip} as a baseline for VCIL. Note that this baseline is trained on images and evaluated in a zero-shot manner, so it requires no fine-tuning on the sequence of video tasks. We denote this as CLIP in Table~\ref{tab:main_results}. Despite not being trained on video data, the CLIP baseline significantly outperforms the memory-based methods iCaRL and BiC reported in the vCLIMB benchmark. The performance gap is especially large in the 20-task setups, where the accuracy of CLIP outperforms iCaRL by $20\%$ on both Kinetics and ActivityNet. This confirms the importance of having models trained on a large volume of data, even when the domains are different, and the relevance of the MCL that allows using it in a zero-shot manner. 

\begin{table}[t]
\caption{
\textbf{Results on UCF101.} We report the average accuracy (Acc) and the backward forgetting (BWF) at 10 and 20 tasks. It is important to note that in this benchmark, the CLIP zero-shot does not outperform the baselines proposed in \cite{Villa_2022_CVPR}. However, our PIVOT model outperforms the previous baselines with great consistency through the number of tasks. 
}
\centering
 \small
\resizebox{8cm}{!}{
\begin{tabular}{cccccc} 
    \toprule

    \multicolumn{1}{c}{\multirow{4}{*}{\bf Model}} & \multicolumn{1}{c}{\multirow{3}{*}{\begin{tabular}[c]{@{}c@{}}\textbf{Mem. Video} \\ \textbf{Instances} \end{tabular}}} & \multicolumn{4}{c}{\bf UCF101}
    \\\cmidrule(lr){3-6}
         & & \multicolumn{2}{c}{\bf 10 Tasks} & \multicolumn{2}{c}{\bf 20 Tasks}
         \\\cmidrule(lr){3-6}
         & & \multicolumn{1}{c}{\bf Acc $\uparrow$ } & \multicolumn{1}{c}{\bf BWF $\downarrow$ } & \multicolumn{1}{c}{\bf Acc $\uparrow$ } & \multicolumn{1}{c}{\bf BWF $\downarrow$ }
         \\
    \midrule
    EWC & None & 9.51\% & 98.94\% & 4.71\% & 92.12\%
    \\
    MAS & None & 10.89\% & 11.11\% & 5.90\% & 5.31\%
    \\
    \cmidrule{1-6}
    BiC & 2020 & 78.16\% & 18.49\% & 70.69\% & 24.90\%
    \\
    iCaRL & 2020 & 80.97\% & 18.11\% & 76.59\% & 21.83\%
    \\
    \cmidrule{1-6}
    CLIP & None & 65.73\% & 11.14\% & 65.38\% & 12.35\%
    \\
    \cmidrule{1-6}
    PIVOT  & 1010 & 93.36\% & 4.47\% & 93.07\% & \bf 3.90\% 
    \\
    PIVOT w/o prompts & 1010 & \bf 94.80\% & \bf 3.89\% & \bf 93.70\% & 4.77\%
    \\
    \cmidrule{1-6}
    Upper bound & -- & 94.9\% & -- & 94.9\% & --
    \\
    \bottomrule
    \end{tabular}
}
\vspace{-4mm}
\label{tab:ucf_results}
\end{table}

\vspace{2pt}\noindent\textbf{The Effect of Long CIL Sequences.} Our results in Table~\ref{tab:main_results} show that a frozen image-pretrained model does not experience extra forgetting when the length of the CIL sequence increases. Moving from 10 to 20 tasks in Kinetics does not affect CLIP's performance as it only changes from $46.5\%$ to $46.52\%$. Similarly, for ActivityNet, the accuracy values of CLIP barely differ, going from $64.1\%$ to $64.13\%$. This makes sense as the full clip encoder is frozen, so presenting the action categories in a shorter or longer sequence does not change the model's behavior along the task set. 

Likewise, the last two rows of Table~\ref{tab:main_results}, show that PIVOT and PIVOT w/o prompts significantly improve the CLIP baseline performance without resulting in a performance gap between the 10-task and 20-task sequences. In both Kinetics and ActivityNet, our models PIVOT and PIVOT w/o prompts are consistent regardless of the task sequence length. They present at most 1.21\% between the accuracy in the two scenarios, which is much less than the 5.43\% presented on \cite{Villa_2022_CVPR}.

\vspace{2pt}\noindent\textbf{Results on UCF101.}
We verify the effectiveness of PIVOT by comparing its performance to SOTA methods on the UCF101 dataset. While iCaRL and BiC perform reasonably well in the 10-task CIL scenario on this less challenging dataset, the final accuracy of frozen CLIP on UCF is around $15,24\%$ behind them, suggesting that this baseline may not always beat traditional CIL approaches. Yet, by effectively leveraging the extensive image pre-training through temporally aware MCL and prompting, PIVOT results in a significant improvement of $28\%$ in accuracy over the CLIP baseline. Consistent with our results on Kinetics and ActivityNet, both PIVOT and PIVOT w/o prompts outperform the methods in vCLIMB with up to $14\%$ improvement over iCaRL, and exhibit a minimal drop in performance between the 10 task and 20 task scenario.  

\vspace{2pt}\noindent\textbf{Comparison with the Fully-supervised Upper Bound.} We would like to highlight from Table~\ref{tab:main_results} and Table~\ref{tab:ucf_results} that the significant improvements, acquired by using PIVOT or PIVOT w/o prompts, result in a final CIL performance that is comparable to the upper bound of training on the whole data offline. While there is a significant gap to the upper bound performance on Kinetics, the gap to the upper bound in the ActivityNet 20-task setting is decreased from around $45\%$ to $15\%$. Moreover, the accuracy of $94.8\%$ obtained by PIVOT w/o prompts in the 10-task UCF101 benchmark already matches the upper bound performance of $94.9\%$.

\begin{table*}[t]
\caption{
\textbf{Ablation study results}. We analyze the impact of each component of PIVOT: the memory buffer, the temporal encoder, the Multimodal Contrastive Loss (MCL), and the prompting approach. For the prompting approach, we compare PIVOT with an extension of L2P for the video domain using CLIP as the backbone (Spatial Prompting). As can be observed, every component of our model contributes significantly to its final performance, which outperforms all the state-of-the-art approaches.}\vspace{-1mm}
\centering
 \small
\resizebox{17cm}{!}{
\begin{tabular}{l ccccccccc} 
    \toprule

    \multicolumn{1}{c}{\multirow{4}{*}{\bf Method}} & \multicolumn{1}{c}{\multirow{4}{*}{\begin{tabular}[c]{@{}c@{}}\textbf{Uses Memory} \end{tabular}}} & \multicolumn{4}{c}{\bf UCF101} & \multicolumn{4}{c}{\bf ActivityNet-Trim}
    \\\cmidrule(lr){3-10}
         & & \multicolumn{2}{c}{\bf 10 Task} & \multicolumn{2}{c}{\bf 20 Task} & \multicolumn{2}{c}{\bf 10 Task} & \multicolumn{2}{c}{\bf 20 Task}
         \\\cmidrule(lr){3-10}
         & & \multicolumn{1}{c}{\bf Acc $\uparrow$ } & \multicolumn{1}{c}{\bf BWF $\downarrow$ } & \multicolumn{1}{c}{\bf Acc $\uparrow$ } & \multicolumn{1}{c}{\bf BWF $\downarrow$} & \multicolumn{1}{c}{\bf Acc $\uparrow$ } & \multicolumn{1}{c}{\bf BWF $\downarrow$ } & \multicolumn{1}{c}{\bf Acc $\uparrow$ } & \multicolumn{1}{c}{\bf BWF $\downarrow$}
         \\
    \midrule
    CLIP Baseline & \xmark & 65.73\% & 11.14\% & 65.38\% & 12.35\% & 64.1\% & 10.47\% & 64.13\% & 11.71\%
    \\
    \cmidrule{1-10}
    + Spatial Prompting & \xmark & 77.66\% & 9.79\% & 68.66\% & 15.19\% & 64.21\% & 12.57\% & 57.89\% & 15.14\%
    \\
    \cmidrule{1-10}
    + Memory Buffer & \cmark & 88.42\% & 11.91\% & 87.52\% & 11.94\% & 63.33\% & 26.17\% & 62.10\% & 28.29\%
    \\
    \cmidrule{1-10}
    + Classification with MCL & \cmark & 86.17\% & 10.65\% & 87.59\% & 9.05\% & 69.53\% & 12.44\% & 69.32\% & 13.51\% \\
    \cmidrule{1-10}
    + Temporal encoder & \cmark & 91.09\% & 5.95\% & 90.04\% & 5.19\% & 70.37\% & 10.83\% & 72.15\% & 10.02\%
    \\
    \cmidrule{1-10}
    + improved prompting \textbf{(PIVOT)}  & \cmark & 93.36\% & 4.47\% & \bf 93.07\% & \bf 3.90\% & \bf 73.8\% & \bf 11.41\% & \bf 73.84\% & \bf 11.94\%
    \\
    \textbf{PIVOT w/o prompts}  & \cmark & \bf 94.80\% & \bf 3.89\% & 93.70\% & 4.77\% & 72.22\% & 13.71\% & 73.43\% & 13.45\%
    \\
    \bottomrule
    \end{tabular}
}
\vspace{-3mm}
\label{tab:ablation}
\end{table*}

\subsection{Relevance of each component of PIVOT}  \label{sec:ablation_study}
To demostrate the effectiveness and contribution of each component of PIVOT, we provide a series of experiments on UCF101 and ActivityNet. We start from the frozen CLIP baseline and we iteratively add one element of PIVOT to test its effect. The results are summarized in Table~\ref{tab:ablation}. 

\vspace{2pt}\noindent\textbf{Effect of spatial prompts.} We first add spatial learnable prompts to modulate the frozen CLIP, and we replace the MCL strategy with a linear classifier. The goal of these prompts is to guide the prediction of the classifier based on task-specific spatial information. We see from Table \ref{tab:ablation} these spatial prompts are useful in UCF101, resulting in a $12\%$ increase in accuracy in the 10-task scenario. However, spatial prompting alone is not useful in ActivityNet. This can be explained by the complicated temporal dependencies that are found in ActivityNet but not in UCF101. 
We show later that augmenting these spatial prompts with temporal prompts can increase the performance in ActivityNet.

\vspace{2pt}\noindent\textbf{Effect of memory replay.} We extend the previous method by allowing the model to store a few video samples for future replay. As an incremental analysis, we keep using spatial prompts from the previous. This addition results in a further performance increase of $11\%$ in the 10-task setting of UCF101 and $19\%$ in the 20-task setting. This model is already ahead of the CLIP baseline by around $23\%$. 
\begin{figure}
    \centering
\includegraphics[width=0.8\linewidth]{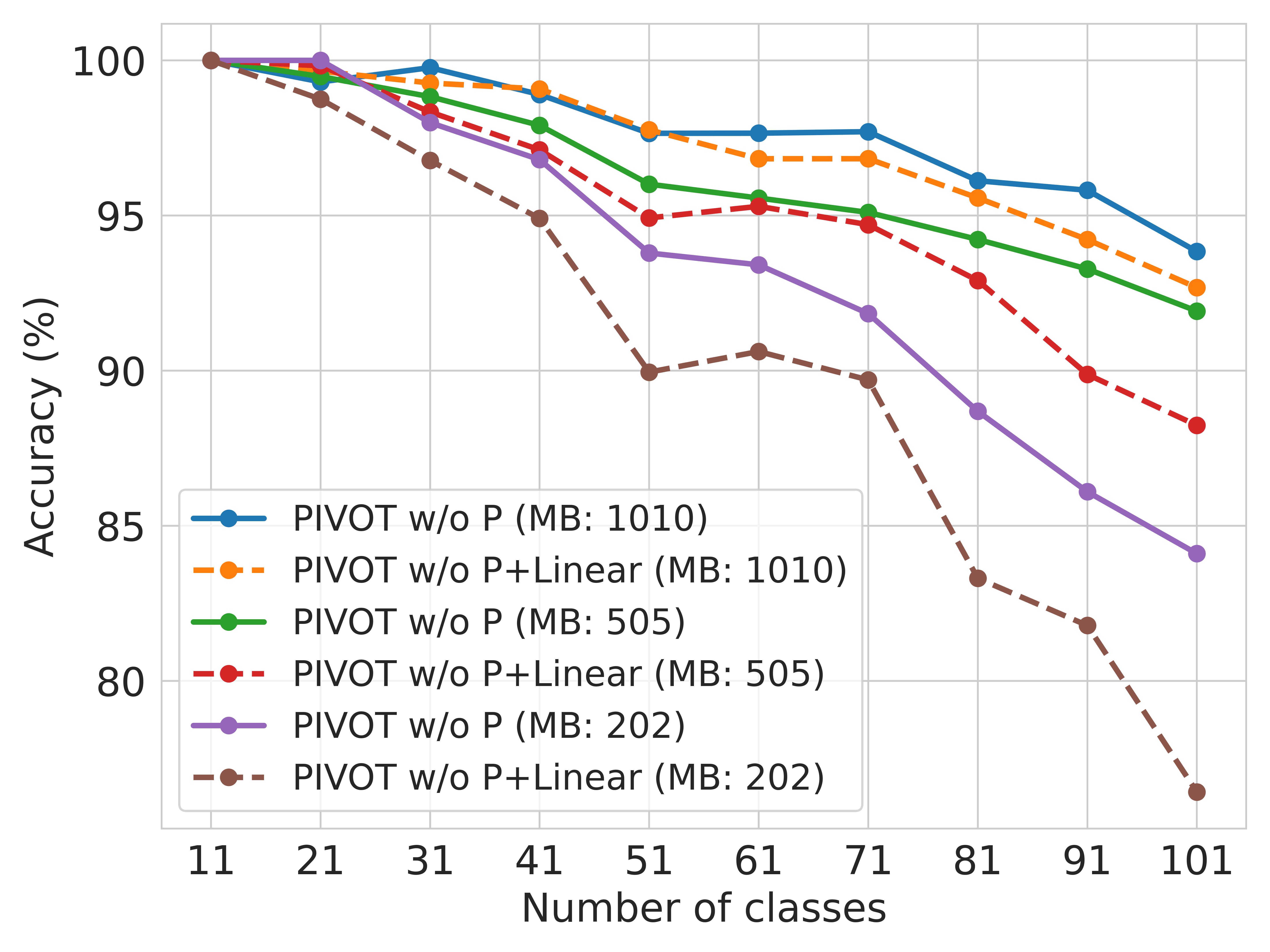}\vspace{-0.2cm}
    \caption{\textbf{MCL robustness to the memory size.} All the experiments were done in the 10-task scenario of UCF101 considering different memory sizes (1010, 505, and 202) and two types of classifiers (Linear and MCL) for our PIVOT w/o prompts model.
    }\vspace{-0.3cm}
    \label{fig:MCL_effect}
\end{figure}

\vspace{2pt}\noindent\textbf{Effect of the Multimodal Contrastive Classifier.} To address the temporal variation, we employ the MCL strategy explained in \ref{sec:mcl}. Table \ref{tab:ablation} shows that this addition results in a $5\%$ increase in final accuracy over the CLIP baseline in the challenging ActivityNet dataset. Moreover, as observed in Figure~\ref{fig:MCL_effect}, the MCL is more robust to the number of instances in the memory than the traditional linear classifier.

\vspace{2pt}\noindent\textbf{Effect of the Temporal Modeling.} 
As one of the limitation for using CLIP in videos context is the lack of temporal modeling in it architecture.
To improve this point in VCIL, we utilize a transformer encoder to model the temporal information of the input videos. This encoder further mitigates forgetting, decreasing the BWF from $10.65\%$ to $5.95\%$ in the 10-task UCF101 benchmark.

\vspace{2pt}\noindent\textbf{Effect of the improved Prompts.} Given the transformer encoder, we can learn the temporal prompts. The resulting model is PIVOT, which further improves the accuracy on ActivityNet by around $3.5\%$. Finally, we show that PIVOT w/o prompts, which removes the prompting strategy from this model but keeps all the other components, results in the best performance on UCF101.

%% file: sections/5_conclusion.tex
\section{Conclusion and Limitations}
\label{sec:conclusion}
We proposed PIVOT, a novel method that leverages large-scale pretraining from the image domain to mitigate catastrophic forgetting in video class incremental learning. 
PIVOT is the first prompting-based approach for video continual learning. While requiring much less memory than existing baselines, PIVOT outperforms these baselines by $31\%$, $27\%$, and $17.2\%$ in the 20-task setups of the vCLIMB benchmark. 
Despite PIVOT's advantages, we found it difficult to fine-tune the CLIP encoder for VCIL.
PIVOT overcomes this problem by leveraging learnable prompts and a parameter-free multimodal contrastive classifier. 
However, we believe that recent advances in robustly fine-tuning large pre-trained models under distribution shifts~\cite{wortsman2022robust} could lead to even greater improvements for PIVOT.

\vspace{2pt}\noindent\textbf{Acknowledgments.} This work was supported by the King Abdullah University of Science and Technology (KAUST) Office of Sponsored Research (OSR) under Award No. OSR-CRG2021-4648, as well as the SDAIA-KAUST Center of Excellence in Data Science and Artificial Intelligence (SDAIA-KAUST AI). Likewise, it was also partially funded by FONDECYT grant 1221425 and the National Center for Artificial Intelligence CENIA FB210017, Basal ANID.

%% file: sections/appendix.tex
\onecolumn
\section{Prompt Selector}\label{sec:supp_prompt_selector}
\begin{figure*}[t]
    \centering
    \includegraphics[width=1.0\linewidth]{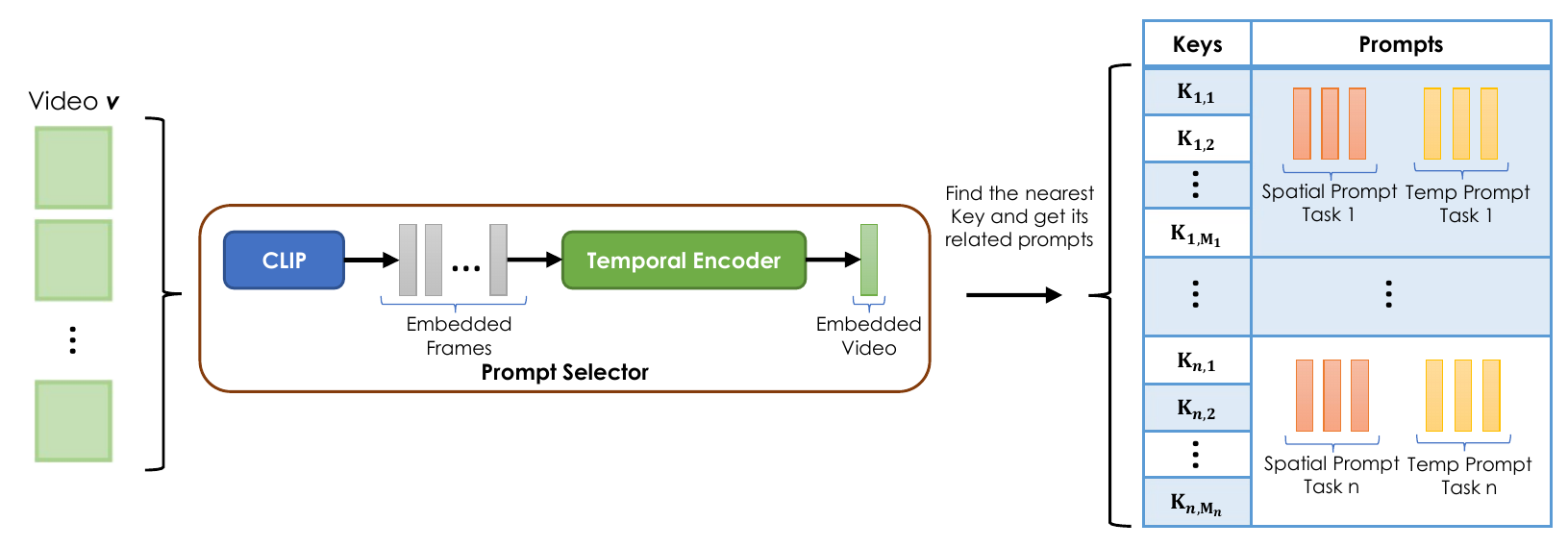}
    \caption{\textbf{Outline of Prompt Selector.} Our prompt selector leverages the CLIP $f_{sp}$ and Temporal Encoder $f_{tp}$ as the query function $q = f_{sp} \circ f_{tp}$ to select the suitable task prompts based on the similarity between the encoded video $q(\textit{\textbf{v}})$ and the keys $K$ of all task prompts. 
    }
    \label{fig:prompt_selector}
\end{figure*}

In Section~\ref{sec:pivot_model}, we outlined our key components in PIVOT. The Prompt Selector module in PIVOT selects the most suitable prompts for classifying a given input. 
Here we delve into the details of our Prompt Selector module.

Given a class incremental sequence of $n$ tasks, PIVOT learns a set of $n$ task-specific prompts $P_i=\{(P_i^{sp}, P_i^{tp})\}_{i=1}^n$, where each prompt $P_i$ is a key-value pair $(K_i, P_i)$. Note that $K_i$ consists of the encoded representations of the $M_i$ labels found in task $i$. These representations are obtained using the CLIP Text encoder. As can be seen in Figure \ref{fig:prompt_selector}, our prompt selector passes the video $\textit{\textbf{v}}$ through the CLIP visual encoder $f_{sp}$ and our temporal encoder $f_{tp}$ to compute a query $f_{tp}(f_{sp}(\textit{\textbf{v}}))$. Following the tokenization procedure of CLIP, each input representation includes a [class] token. The computation of our query takes into account the [class] tokens encoded in the representations of both the visual and temporal encoders. PIVOT selects the corresponding task prompts for $\textit{\textbf{v}}$ based on the similarity between the query and all keys. 

We present Algorithm~\ref{alg:pivot_v2_forward_pass}, which summarizes the prompt selection process of PIVOT and how it uses the selected prompts. Likewise, Algorithm~\ref{alg:pivot_forward_pass} clarifies the forward pass of our base PIVOT w/o prompts, which does not require the task-specific prompts.

\RestyleAlgo{ruled}

\SetKwComment{Comment}{/* }{ */}

\begin{algorithm*}[hbt!]
\caption{PIVOT w/o prompts Forward Pass} \label{alg:pivot_forward_pass}
\KwData{}
$\textit{\textbf{Y}} = (\textit{\textbf{y}}_1, ...,\textit{\textbf{y}}_M)$\ \Comment*[r]{The representations of all learned classes computed with the CLIP Text Encoder.}

\textbf{Components:}

$f_{sp}, f_{tp}$\ \Comment*[r]{CLIP Model, Temporal Encoder}
$\gamma(.,.), f_{cls}$\ \Comment*[r]{Cosine distance function, Select the class label whose representation is most similar to the video}

\textbf{Forward Pass:}

$\textit{\textbf{v}}_{tp} = f_{tp}(f_{sp}(\textbf{\textit{v}}))$ where $\textbf{\textit{v}} \in D_t$\ \Comment*[r]{Compute the video
embedding}
$y = f_{cls}(\textit{\textbf{v}}_{tp},\textit{\textbf{Y}})$\ \Comment*[r]{Classify the video}

\end{algorithm*}

\section{Number of Trainable Parameters in PIVOT} Considering the implementation details presented in section \ref{sec:results}, we analyze and compare the trainable parameters of our PIVOT model against the baseline models. Note that the vCLIMB baselines utilize TSN as a backbone and train it at every task. On the other hand in PIVOT w/o prompts, we leverage the extensive knowledge of CLIP by freezing its visual and text encoders, so we only train our temporal encoder. In PIVOT, we further learn task-specific prompts, which does not result in a significance increase in the number of parameters. As a result of leveraging the knowledge in CLIP without fine-tuning it, we substantially reduce the number of total trainable parameters. Table \ref{tab:parameters} shows that PIVOT w/o prompts and PIVOT train at most 40.56\% of the parameters that the vCLIMB baselines train. It is worth highlighting that the task-specific prompts correspond to an increase of 0.40\% and 0.80\% of PIVOT w/o prompts parameters in the 10-task and 20-task scenarios, respectively. Thus, the resulting PIVOT is comparable in the number of parameters to PIVOT w/o prompts. 

\begin{table}[t]
\caption{
\textbf{The number of trainable parameters.} All the vCLIMB Baselines use the TSN with ResNet50 as a backbone, in addition to a linear layer to perform the classification. These models have approximately the same number of parameters. We consider ActivityNet, which have 200 classes in total, to compute the number of parameters of the linear layers. We note that PIVOT w/o prompts and PIVOT results in an order of magnitude reduction in the number of parameters to train in video class incremental learning. 
}
\centering
 \small
\begin{tabular}{cc} 
    \toprule
    \multicolumn{1}{c}{Model} & \multicolumn{1}{c}{Num. Trainable Parameters} 
    \\\midrule
    vCLIMB Baselines & $23.610632 \times 10^6$
    \\
    \cmidrule{1-2}
    PIVOT (10-task) & $9.496064\times 10^6$
    \\
    PIVOT (20-task) & $9.534464\times 10^6$
    \\
    PIVOT w/o prompts & $9.457664\times 10^6$
    \\
    \bottomrule
    \end{tabular}

\label{tab:parameters}
\end{table}

\section{Prompt Hyper-parameter Analysis}  
As observed in Table~\ref{tab:hp_parameters}, we explored different configurations for the prompt length $L$ and number of prompts $N_p$ per task on the validation set of the most challenging dataset we evaluated, ActivityNet. We considered the same $L$ for both spatial and temporal prompts $(L=L_{sp}=L_{tp})$. 
PIVOT is more sensitive to the number of prompts per task than their length. It is important to note that $N_p=1$ and $L=3$ for both spatial and temporal prompts work better for ActivityNet. For simplicity, we use the same setup for the other datasets we evaluated.

\begin{table}[t]
\caption{\textbf{Prompt Hyper-parameters.} We vary the prompt length $(L)$ and number of prompts per task $(N_p)$ and report PIVOT performance. To assess $L$, we fix $N_p=1$ and use the same $L$ for both spatial and temporal prompts $(L = L_{sp} = L_{tp})$. Likewise, for $N_p$, we set $L$ to its optimal value $(L=3)$. 
}

\centering
 \small
\begin{tabular}{cc|cc} 
    \toprule
    \multicolumn{1}{c}{Length of prompts} & \multicolumn{1}{c|}{Acc} &
    \multicolumn{1}{c}{Num. Prompts} & \multicolumn{1}{c}{Acc} 
    \\\midrule
    PIVOT $(L=1)$ & 73.2\% & PIVOT $(N_p=1)$ & \bf 73.8\%
    \\
    PIVOT $(L=3)$ & \bf 73.8\% & PIVOT $(N_p=2)$ & 72.60\%
    \\
    PIVOT $(L=5)$ & 73.1\% & PIVOT $(N_p=4)$ & 70.59\%
    \\
    PIVOT $(L=7)$ & 72.7\% & PIVOT $(N_p=6)$ & 69.11\%
    \\
    \bottomrule
    \end{tabular}

\vspace{-6mm}
\label{tab:hp_parameters}
\end{table}


\begin{algorithm*}[hbt!]
\caption{PIVOT Forward Pass} \label{alg:pivot_v2_forward_pass}
\KwData{}
$\textit{\textbf{Y}} = (\textit{\textbf{y}}_1, ...,\textit{\textbf{y}}_M)$\ \Comment*[r]{The representations of all learned classes computed with the CLIP Text Encoder}

\textbf{Components:}

$f_{sp}, f_{tp}$\ \Comment*[r]{CLIP Model, Temporal Encoder}
$f_{sp} = f_{sp}^{e} \circ
 f_{sp}^{s}$'\ \Comment*[r]{Where $f_{sp}^{e}$ is the input layer and $f_{sp}^{s}$ the self-attention layers}
$f^{avg}_{sp}, f^{avg}_{tp}$\ \Comment*[r]{Compute an average pooling through the added spatial and temporal prompts}
$(K_i, P_i)_{i=1}^n$\ \Comment*[r]{Set of prompts for the n tasks}
$K_i \in \mathbb{R}^{M_i \times D_m}$\ \Comment*[r]{Keys of task $i$, where $M_i$ is the number classes of task $i$}
$P_i = (P^{sp}_i, P^{tp}_i)$\ \Comment*[r]{Spatial and Temporal Prompts of task $i$}
$\gamma(.,.), f_{cls}$\ \Comment*[r]{Cosine distance function, Select the class label whose representation is most similar to the video}

\textbf{Forward Pass:}

$\textit{\textbf{v}}_{tp} = f_{tp}(f_{sp}(\textbf{\textit{v}}))$ where $\textbf{\textit{v}} \in D_t$\ \Comment*[r]{Compute the Query}
$P_k = f_{min}( \gamma(\textit{\textbf{v}}_{tp}, K))$\ \Comment*[r]{Select the task prompt that is closet the Query}
$\textit{\textbf{v}}^{e}_{sp} = [P_k^{sp}; f^{e}_{sp}(\textbf{\textit{v}})]$\ \Comment*[r]{Add the spatial prompt}
$\textit{\textbf{v}}^{avg}_{sp} = f^{avg}_{sp}(f^{s}_{sp}(\textit{\textbf{v}}^{e}_{sp}))$\ \Comment*[r]{Compute the representation of the frames per video}
$\textit{\textbf{v}}_{tp} = f_{tp}^{avg}(f_{tp}([P^{tp}_k, \textit{\textbf{v}}^{avg}_{sp}]))$\ \Comment*[r]{Add the temporal prompt and Compute the final video embedding}
$y = f_{cls}(\textit{\textbf{v}}_{tp},\textit{\textbf{Y}})$\ \Comment*[r]{Classify the video}
\end{algorithm*}